# AI paradigm for solving differential equations: first-principles data generation and scale-dilation operator AI solver

Xiangshu Gong[1,†], Zhiqiang Xie[1,†], Xiaowei Jin[1,†,*], Chen Wang[1], Yanling Qu[2], Wangmeng Zuo[3,*], Hui Li[1,3,*]

[1] Center for Intelligent Fluid Mechanics, Harbin Institute of Technology, Harbin, China
[2] Institute of Engineering Mechanics, China Earthquake Administration, Harbin, China
[3] School of Computer Science, Harbin Institute of Technology, Harbin, China
[†] These authors contributed equally to this work.
[*] Corresponding authors: xiaowei.jin@hit.edu.cn, wmzuo@hit.edu.cn, lihui@hit.edu.cn

Many problems are governed by differential equations (DEs). Artificial intelligence (AI) is a new path for solving DEs. However, data is very scarce and existing AI solvers struggle with approximation of high frequency components (AHFC). We propose an AI paradigm for solving diverse DEs, including DE-ruled first-principles data generation methodology and scale-dilation operator (SDO) AI solver. Using either prior knowledge or random fields, we generate solutions and then substitute them into the DEs to derive the sources and initial/boundary conditions through balancing DEs, thus producing arbitrarily vast amount of, first-principles-consistent training datasets at extremely low computational cost. We introduce a reversible SDO that leverages the Fourier transform of the multiscale solutions to fix AHFC, and design a spatiotemporally coupled, attention-based Transformer AI solver of DEs with SDO. An upper bound on the Hessian condition number of the loss function is proven to be proportional to the squared 2-norm of the solution gradient, revealing that SDO yields a smoother loss landscape, consequently fixing AHFC with efficient training. Extensive tests on diverse DEs demonstrate that our AI paradigm achieves consistently superior accuracy over state-of-the-art methods. This work makes AI solver of DEs to be truly usable in broad nature and engineering fields.

Behaviors of our planets, natural world and man-made materials and systems are governed by differential equations (DEs), including ordinary differential equations (ODEs) and partial differential equations (PDEs). Solutions of DE are the pathway to understand their intrinsic laws and mechanism. Obtaining solutions to DE is a central pursuit for scientists engaged in research and engineers involved in design across a wide range of disciplines. However, traditional DE solvers, e.g., finite difference/elementary/volume methods, can only compute a single solution for each given set of conditions. Moreover, for multiscale problems, the high computational cost of temporal and spatial discretization often renders the problem intractable, due to nonlinearity and the curse of dimensionality. Advantageously, neural networks can overcome the curse of dimensionality in function approximation, thereby providing a viable pathway to addressing high-dimensional challenges characterized by multi-scale and many-degree-of-freedom systems[1].

Recently, Artificial Intelligence (AI) provides a potential tool to fix this challenge[2]. Early machine learning methods for solving PDEs include Physics-Informed Neural Networks (PINNs)[3], the Deep Ritz method[4], and the Deep Galerkin method[5], among others. These methods represent PDE solutions as neural network approximations of spatiotemporal functions. They obtain the solutions by embedding equation residuals into the loss function and training neural networks until convergence. Such methods have achieved significant success across various scientific domains, including fluid mechanics[6,7], solid mechanics[8], and aerospace engineering problems[9]. However, these approaches require retraining for each new set of PDE parameters, initial conditions, or boundary conditions, resulting in low computational efficiency. Their primary advantage lies in solving inverse problems, while offering no significant benefit over traditional numerical methods for solving forward problems.

Subsequently, operator learning methods for solving parameterized PDEs emerged, aiming to establish a mapping from the equation's parameter space or initial/boundary condition space to the solution space. Representative methods include DeepONet[10] and Fourier Neural Operator (FNO)[11]. A dual-network architecture consisting of a "BranchNet" and a "TrunkNet" was constructed in DeepONet, where the BranchNet takes as input the function values at selected spatiotemporal points, and the TrunkNet takes the coordinates in the output space. The outputs of these two networks were then combined to predict the solution function at the specified coordinates. The design of DeepONet directly leveraged the universal approximation theorem for operators[12], thus providing theoretical assurance for learning operators with deep neural networks. FNO[11] exploited the low-rank structure of PDE solutions in the spectral domain, incorporating the Fourier transform into the network architecture. By parameterizing integral kernels in the frequency domain, FNO could efficiently learn the mapping from PDE parameters to solutions, achieving high-performance approximations for complex physical problems. A detailed comparison and analysis of DeepONet and FNO can be found in Ref[13]. Subsequent improvements have emerged based on DeepONet and FNO. For instance, a DeepONet variant using a residual U-Net as the TrunkNet was proposed, achieving high-accuracy predictions across multiple scales[14]. Jin et al.[15] proposed sub-operator enhanced neural networks, which first learn individual subsystems within complex PDEs and then integrate the learned sub-operators to solve the original equation. As for FNO, Factorized FNO decomposed the Fourier transform and shared integral kernel operators, reducing the number of trainable parameters and attaining higher accuracy than the original FNO in solving the Navier–Stokes equations[16]. Geo-FNO learned geometric deformations in the latent space, addressing the applicability of FNO to irregular geometries[17]. Rahman et al.[18] developed a U-shaped neural operator (U-NO), in which the inner integral operator is implemented based on the Fourier transform-based integration method developed in FNO, allowing for deeper neural operators. To address transient response and non-periodic signal problems, Cao et al.[19] proposed the Laplace Neural Operator (LNO), which overcomes the limitations of the FNO; this network leverages the advantages of the Laplace transform to decompose the input space, and designs learnable pole and residue parameters in the Laplace domain, enabling the construction of a more physically meaningful and interpretable mapping. More recently, methods leverage the mainstream Transformer architecture for solving PDEs are developed by researchers. For instance, Transolver introduced a "physics-attention" mechanism[20], which adaptively grouped point with similar physical states into learnable 'slices'; these slices were then aggregated into "physics-aware tokens" enabling efficient attention computation. This shifted the focus of learning from complex geometric domains to the more essential domain of physical states. PDEformer integrated attention mechanisms with graph-based representations to solve partial differential equations across diverse disciplines[21]. Wu et al. proposed the latent spectral models (LSM) to solve PDEs in the latent space by learning multiple basis operators[22]. Although the above AI solvers have achieved progress in various aspects of solving differential equations compared to traditional numerical methods, training these methods often requires a large amount of equation data. Such data typically needs to be generated using conventional numerical solvers or experimental approaches, which demand substantial resources, making data scarcity a prominent issue. Researchers have become aware of the data scarcity problem in training large models and have begun leveraging foundation models to synthesize data, e.g., images[23]. However, obtaining arbitrary amounts of scientific computing data for training large equation-solving models remains highly challenging. Furthermore, these AI methods still face significant challenges in solving multiscale problems, particularly in accurate approximation of high frequency components (AHFC)[24,25].

To this end, we first propose a DE-ruled data generation method (Fig. 1) for training AI perspective DE solvers. A general form of a DE is written as

$$\mathcal{L}_t(u(t,x);\alpha) + \mathcal{L}_x(u(t,x);\alpha) = f(t,x), \quad (t,x) \in [0,T] \times \Omega, \tag{1a}$$
$$u(0,x) = u_0(x), \quad x \in \Omega; u(t,x) = g(t,x), \quad x \in \partial\Omega, \tag{1b}$$

where $u(t,x) \in \mathbb{R}^{d \times 1}$ is to be solved function of a physical quantity, defined on a spatial domain $\Omega$ over the time interval $[0,T]$. The $\mathcal{L}_t$ and $\mathcal{L}_x$ are time derivative operator and spatial differential operator. The $f(t,x)$ is source function, $u_0(x)$ and $g(t,x)$ are the initial condition and boundary condition on $\partial\Omega$. $\alpha$ is DE's parametric set controlling its intrinsic behavior.

Typically, numerical methods solve $u(t,x)$ given initial and boundary conditions, and source with a numerical scheme, $(u_0(x), g(t,x), f(t,x)) \mapsto u(t,x)$. Usually, $u(t,x)$ is a complex nonlinear function in high-dimensional space, resulting in very expensive numerical simulation. In contrast, we first generate $u(t,x)$ either randomly or following prior knowledge, readily calculate $\mathcal{L}_t$ and $\mathcal{L}_x$, then naturally derive $u_0(x), g(t,x), f(t,x)$ through balancing the DE, $u(t,x) \mapsto (u_0(x), g(t,x), f(t,x))$. As a consequence, we get arbitrary large amount of data pairs $\{(u_0(x), g(t,x), f(t,x)), u(t,x)\}$ at very low cost and high efficiency. This method seems greedy, the initial/boundary condition may not exist before, while it does not matter in AI perspective, the vast data pairs just embed the intrinsic mapping relationship from source, initial and boundary conditions to solution of DE. Thus, the data pairs can be used to train a good AI solver of DEs, and AI solver captures nature of DEs can then be used to solve problems in real world.

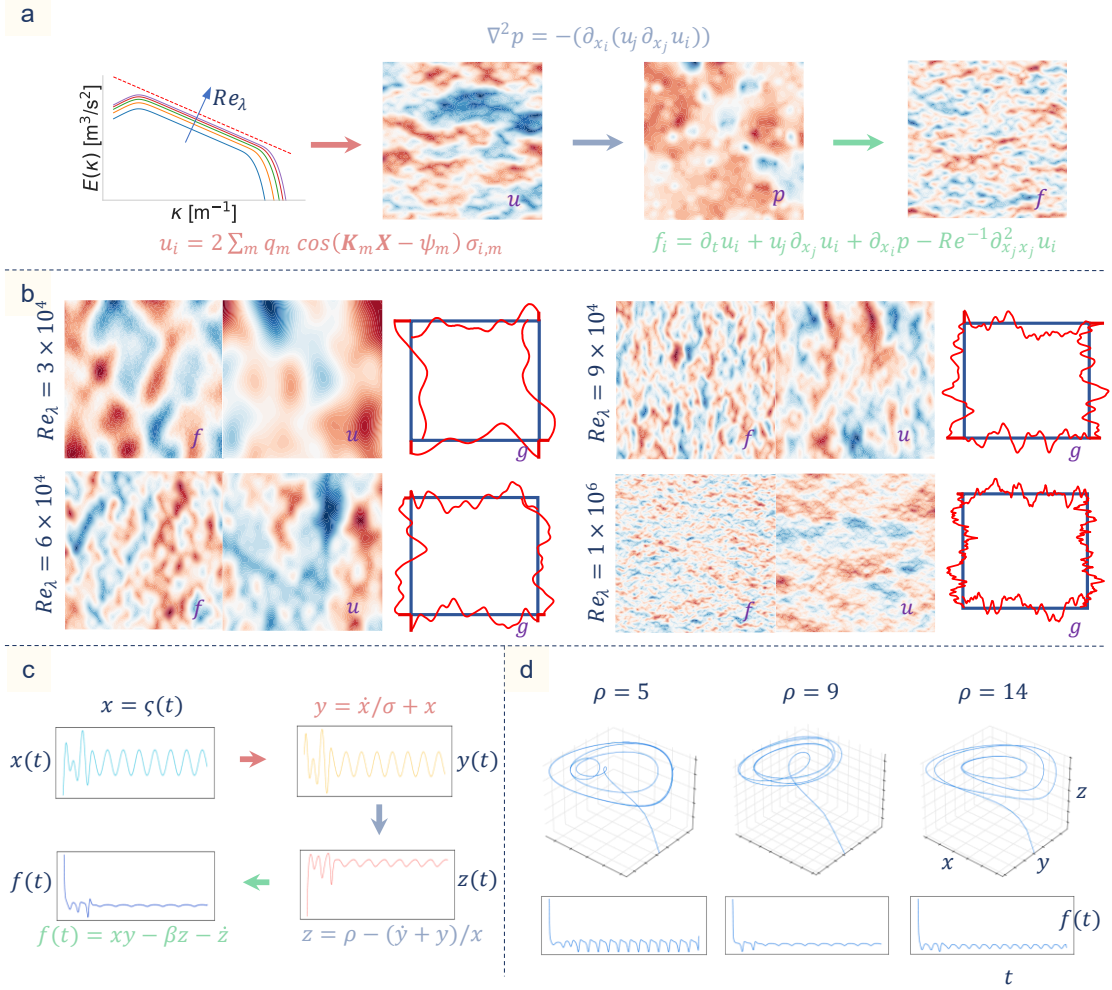

**Fig. 1 | Illustration of DE-ruled data generation method. a**, data generation process of Naiver-Stokes equations. **b,** generated turbulent flows at various Reynolds numbers. **c**, data generation process of the forced Lorenz system. **d,** generated solution of Lorenz system at various Rayleigh numbers.

The solution of DEs can be generated using prior knowledge. For instance, in turbulent flows, spatial energy spectra derived by Pope[26] satisfying Kolmogorov's K41 theory[27] can be incorporated into spatiotemporal energy spectra by Wilczek et al.[28], followed by synthesizing divergence-free $\nabla \cdot u(t, x) = 0$ and energy spectra-consistent solutions in Fourier space. In cases lacking prior models, solutions can alternatively be synthesized using Gaussian random fields as $u(t,x) = \mathcal{F}^{-1}\left\{\sqrt{S(K)\Delta k \Delta \omega} \cdot z(k, \omega)\right\}$, where $k$ and $\omega$ are the wavevector in space domain and frequency in time domain, $z$ is a standard joint Gaussian distribution of $k$ and $\omega$, $S(K)$ is a power spectral density function controlling the correlation structure, and $\mathcal{F}^{-1}$ denotes the $d$-dimensional inverse Fourier transform operator. The time derivative $u_t$ and space gradient $u_x, u_{xx}$ of $u(t,x)$ is calculated, substituting the solution function into DE (1) derives the source and initial/boundary conditions. Because $u(t,x)$ can be arbitrary sampled from spatiotemporal energy spectra, thus we can generate any amount of data pairs of $\{\{u_0(x), g(t,x), f(t,x)\}, u(t,x)\}$ for training AI solver. Navier-Stokes equations for fluids, wave equation, Lorenz equation and equation of motion for dynamics, and Steady Navier–Cauchy equations for solid mechanics are adopted to generate data pairs with various controlling parameters of DEs. We show representative generated data from Navier-Stokes equations and Lorenz equation in Fig. 1. The generated turbulent flow at any high Reynolds number is so attractive because they are very difficult or even impossible to get them through either traditional numerical simulation or experiment. We will present the critical role of generated data pairs for training AI solver in results.

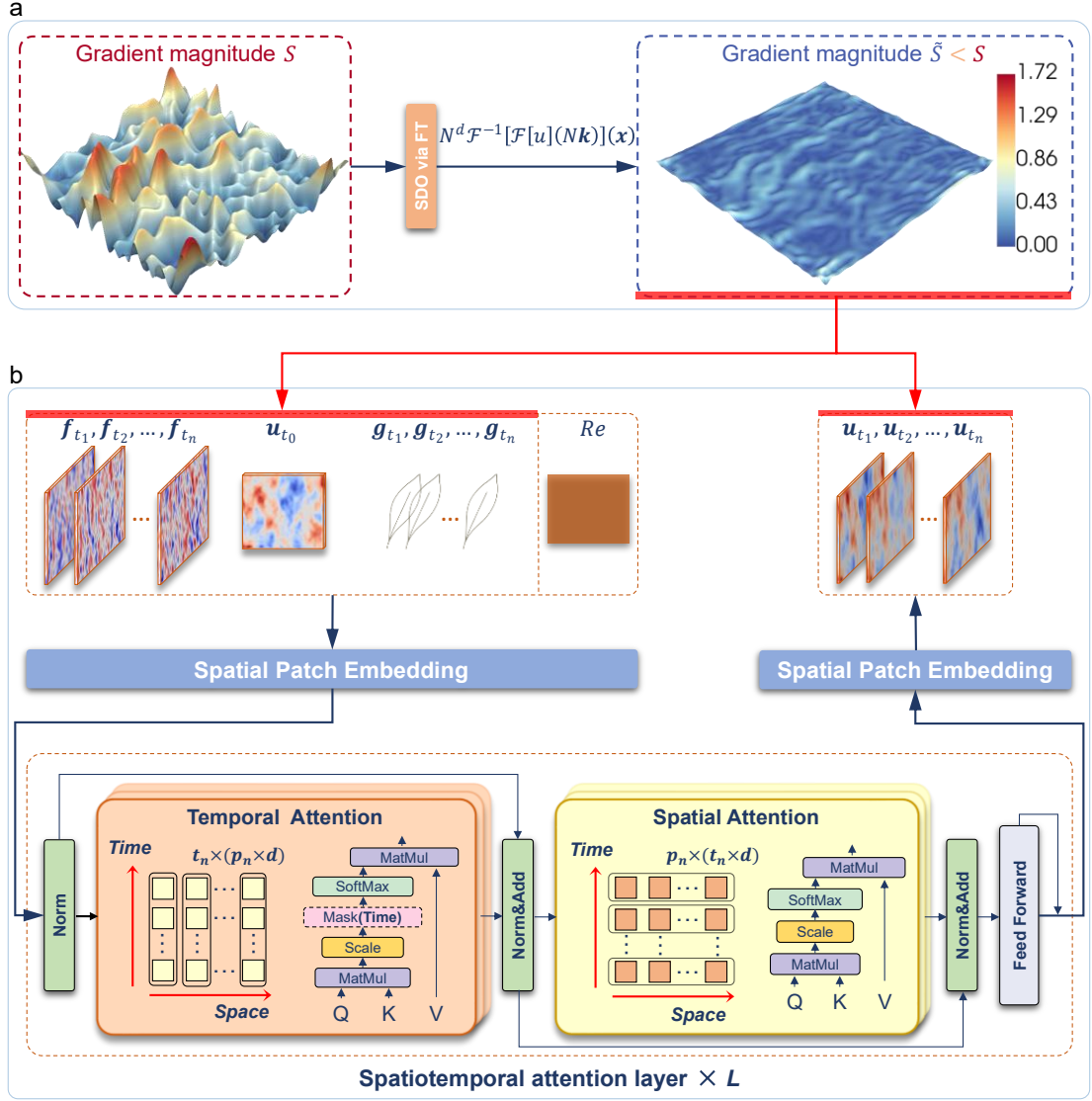

**Fig. 2 | Spatiotemporally coupled, attention-based transformer architecture integrating scale-dilation operator to learn the solution in the low-freqence space. a,** scale-dilation operator. **b,** spatiotemporally coupled, attention-based transformer architecture.

High frequency components either in space domain (wavenumber) or time domain (frequency) are important in the DE's solution. However, the deep network is proven to be weak in AHFC[24,25]. To fix this problem for AI solver of DEs, we define a scale-dilation operator (SDO) $(\mathcal{D}_N)(\boldsymbol{x}) = u \circ \sigma_N(\boldsymbol{x}) = u(\frac{x}{N}) = N^d \mathcal{F}^{-1}[\mathcal{F}[u](N\boldsymbol{k})](\boldsymbol{x})$ in function spaces via the Fourier transform, $\sigma_N: \mathbb{R}^d \to \mathbb{R}^d$ is linear automorphism (an invertible linear map from $\mathbb{R}^d$ to itself), $N \in \mathbb{N}$ is a scaling dilation factor. We have the following Claim with $\tilde{u}$ being the scale-dilation of $u$

Claim. $$\tilde{u}(\boldsymbol{x}) = N^d \mathcal{F}^{-1}[\mathcal{F}[u](N\boldsymbol{k})](\boldsymbol{x}) = u\left(\frac{x}{N}\right)$$

SDO transforms the high frequency components into the low frequency ones, thereby significantly enhancing the AI solver's training efficiency.

We develop an AI solver with a spatiotemporally coupled attention-based transformer architecture incorporating the SDO (Fig. 2), characterized as a three-level composite operator $\hat{u}_{t_n} = \psi \circ \varphi \circ \phi(f, u_{t_0}, g)$, nonlinear encoder $\phi$ mapping the original physical quantities into a high-dimensional latent space with trainable parameter $w_\phi$, $\varphi$ the spatiotemporal attention layer operator with trainable parameter $w_\varphi$, and linear decoder $\psi$ transform the latent representations back to the physical space with trainable parameter $w_\psi$. The input of AI solver includes source $f = \{f(x,t)\}_{t=t_1}^{t_n}$, (temporal interval of $[t_0, t_n]$), initial field $u(x, t_0)$ and boundary condition $g = \{g(x,t)\}_{t=t_1}^{t_n}$ with dilation on them as Claim. The output of the AI solver is the dilated spatiotemporal solution field $\hat{u}_{t_1:t_n} = \{\hat{u}(x,t)\}_{t=t_1}^{t_n}$ and the real solution function is obtained through reverse dilation operation. Norm-1 loss function, $L_1 = \|u - \hat{u}\|_1$, is adopted because of insensitive to data amplitude, thus can accelerating convergence at early training stage. Our proposed AI solver incorporating the SDO is very efficient for training, and the mechanism is proven that the condition number of the Gauss-Newton approximation of the neural network's loss function Hessian matrix bounded (Theorem 1). Our contributions of this work are summarized as follows:

· We propose an equation-ruled big data generation method, enabling the efficient and low-cost generation of arbitrarily large datasets that adhere to first-principles, for training AI solvers.
· We introduce a novel approach for approximating the high-frequency components of DE solutions using a reversible SDO based on the Fourier transform. This operator maps multiscale solution data into a low-wavenumber space with reduced spatial gradient magnitudes. We further provide a theoretical analysis of the Hessian condition number of the loss function, deriving upper bounds in terms of the spatial gradient magnitude of the target solution data.
· A spatiotemporally coupled attention-based transformer architecture, integrated with the SDO, is developed and demonstrates high accuracy in solving multiscale DEs.
· We construct a unified AI solver applicable to a broad class of known equations, capable of generalizing across diverse equation parameters and varying initial and boundary conditions. The effectiveness of the DE-ruled data generation method for learning physically consistent solutions is verified.

# Results

We collect generated data pairs and numerical simulation data together to form dataset for AI solver training. The advantage using two different resource data is to improve robust and generalization of AI solver. To demonstrate the advantage of generated data and AI solver with SDO, we train our deep network, FNO, U-NO, Transolver, and LSM using numerical simulation data (NSD) alone and the entire dataset using NVIDIA H100 GPUs. Our AI solver achieves more accurate result than FNO, U-NO, Transolver, and LSM for both NSD and the entire dataset (Table 1), especially the entire dataset because of more dramatic derivative variance of generated data. The results indicate that by incorporating generated data into the entire dataset, the average relative $L_1$ error for each of the different deep learning models decreased by 4% to 21%. The generalization is one of most critical capability of AI. We use the trained AI solvers to predict solutions of all five kinds of DEs under various conditions. Our AI solver achieve best approximation of solutions at all cases, especially at cases not included in training dataset (Table 1). For instance, our AI solver predict all-scale flow velocity field of flow around a leaf with very high accuracy even our dataset does not contain any leaves in training, meaning that our AI solver can approximate the solutions of conditions never seen before. The results are so exciting that our AI solver can prediction both laminar flow and turbulence flow. We have constructed a unified AI solver that offers a novel approach to a broad class of equations governing intricate planetary phenomena. For all the DEs, both DE-ruled data generated by our proposed method and NSD are included in the training stage. Our proposed method exhibits compelling results for the most challenging multiscale Navier-Stokes equations, with its relative $L_1$ error performance compared against FNO, U-NO, Transolver, and LSM presented in Fig. 3. The relative $L_1$ errors for all other equations we considered using our AI solver are summarized in Table 2.

**Table 1 | Comparison of relative $L_1$ errors for our proposed AI solver versus FNO, U-NO, Transolver, and LSM training with (w/) and without (w/o) generated data. Channel^J represents turbulent channel flow at $Re_\tau \sim 1000$, and AVG is the average error of all cases.**

|               | GenDat | Cylinder | Channel | Channel^J | Airfoil | Leaf | AVG |
|---|---|---|---|---|---|---|---|
| FNO w/o       |        | 25%      | 28%     | 47%       | 4%      | 60%  | 33% |
| FNO w/        | 9%     | 23%      | 19%     | 40%       | 4%      | 41%  | 23% |
| U-NO w/o      |        | 38%      | 26%     | 35%       | 8%      | 54%  | 32% |
| U-NO w/       | 17%    | 22%      | 23%     | 30%       | 7%      | 36%  | 23% |
| Transolver w/o|        | 45%      | 13%     | 53%       | 6%      | 59%  | 35% |
| Transolver w/ | 12%    | 16%      | 15%     | 20%       | 5%      | 17%  | 14% |
| LSM w/o       |        | 25%      | 16%     | 43%       | 6%      | 46%  | 27% |
| LSM w/        | 9%     | 18%      | 14%     | 40%       | 5%      | 25%  | 19% |
| Ours w/o      |        | 16%      | 7%      | 20%       | 4%      | 30%  | 15% |
| **Ours w/**   | **8%** | **11%**  | **7%**  | **18%**   | **4%**  | **16%** | **11%** |

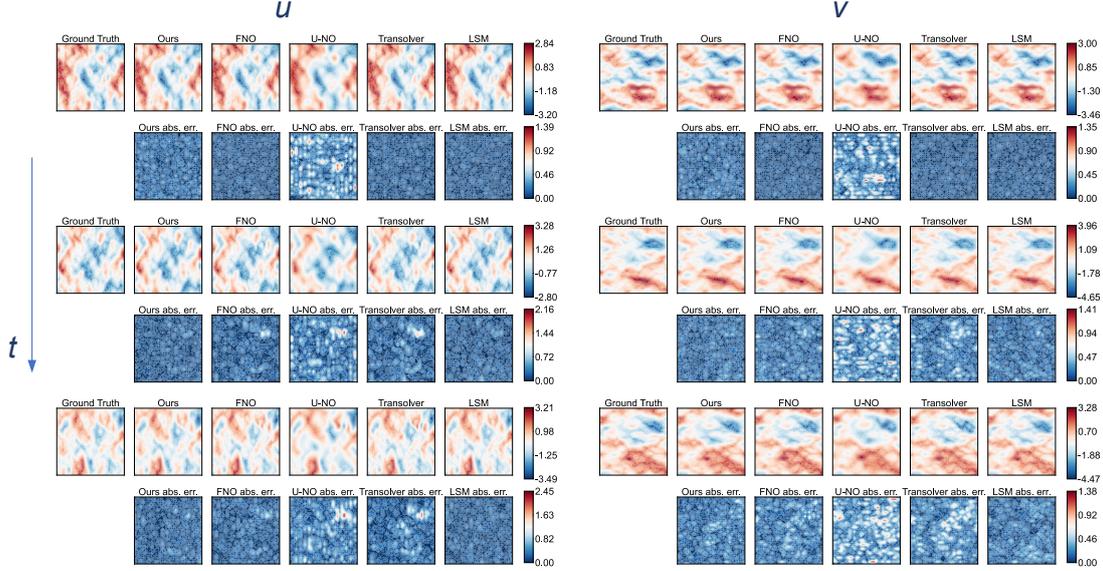

**Fig. 3 | Comparison of solutions for our proposed AI solver versus established methods, including FNO, U-NO, Transolver, and LSM.**

**Table 2 | Relative $L_1$ errors of our proposed AI solvers for various equations.**

| Data set | SNC Eq. | Wave Eq. | EoM | Lorenz Sys. |
|---|---|---|---|---|
| Numerical simulation data | 2% | 6% | 13% | 7% |
| DE-ruled generation data | 2% | 1% | 6% | 3% |

**Navier-Stokes equations**

The Navier-Stokes equations are a set of PDEs that describe the motion of viscous fluid substances, such as water and air. These equations are of fundamental importance in fluid dynamics, used to model systems such as weather, ocean currents, and airflow, and have extensive applications in engineering. The incompressible Navier-Stokes equation is

$$\frac{\partial u_i}{\partial x_i} = 0 \tag{2a}$$

$$\frac{\partial u_i}{\partial t} + u_j \frac{\partial u_i}{\partial x_j} = -\frac{\partial p}{\partial x_i} + \frac{1}{Re}\frac{\partial^2 u_i}{\partial x_j \partial x_j} + f_i \tag{2b}$$

where $u_i$ and $p$ denote the velocity and pressure fields, respectively, $Re$ is the Reynolds number, and $f_i$ is an external forcing term. Eq. (2a) enforces the incompressibility condition, while equation Eq. (2b) governs the momentum dynamics. The system is subject to initial conditions $u(0, x) = u_0(x)$, $x \in \Omega$ and boundary conditions $u(t, x) = g(t, x)$, $x \in \partial\Omega$. In our study, we consider five distinct flow scenarios in the training stage:

- DE-ruled data generation method: velocity, pressure and forcing fields generated using a physically-consistent method that respects the governing equations (shown in Fig. 1).
- Flow behind a circular cylinder: A classical benchmark problem with zero source term $\boldsymbol{f} = \boldsymbol{0}$. This dataset is numerical simulation dataset from Ref[29].
- Channel flow: Featuring a streamwise mean pressure gradient modeled as a nonzero source term. This dataset is numerical simulation dataset via finite difference method at $Re_\tau \sim 110, 150, 180$. Cases with $Re_\tau = 110$ and 180 are used for training, and the case while case $Re_\tau = 150$ is for testing.
- Flow over the S809 airfoil: A standard aerodynamic case with $f = 0$. This dataset is numerical simulation dataset from Ref[30].
- Aneurysmal blood flow with $f = 0$. This dataset is numerical simulation dataset from Ref[31].

Note that flow around a leaf or turbulent channel flow at high Reynolds number $Re_\tau \sim 1000$ are not included in the training dataset, but they are also tested in Table 1. Our objective is to learn the nonlinear mapping defined by Eq. (2): given initial conditions $u_0$ in $\Omega$, boundary conditions $g$ on $\partial\Omega$, the source term $f$ in $\Omega$, and the Reynolds number $Re$, the proposed AI solver predicts the solution field. As demonstrated in Table 1, our proposed AI-solvers achieves high accuracy across a wide range of flow conditions, from relatively simple configurations (e.g., cylinder wake at low Reynolds numbers) to more challenging cases (e.g., DE-ruled data generation flow fields, flow around a leaf, and turbulent channel flow at high Reynolds number $Re_\tau \sim 1000$). Notably, in complex scenarios, the introduction of SDO effectively and accurately simulates high-frequency components in the solution. The SDO results in smoother gradients within the domains of interest and enhances the model's learning capability, outperforming state of the art baselines.

**Steady Navier–Cauchy equations**

The linear elasticity model is governed by a set of second-order linear PDEs that describe the mechanical response of an elastic body subjected to static external loads (i.e., loads that are time-invariant or vary extremely slowly). The model is formulated under the assumptions of linear elasticity, small deformations, and material behavior obeying Hooke's law. For simplification, the analysis is often reduced to either the plane stress or plane strain condition[32], depending on the geometry and boundary constraints of the problem.

For the plane stress problem, by substituting the kinematic relations (strain-displacement equations) and the constitutive equations (stress-strain relations) into the equations, we can obtain the governing equations expressed solely in terms of the displacement components, which are known as the steady Navier–Cauchy equation (SNC) in the context of linear elasticity:

$$\frac{E}{1-\mu^2}\left(\frac{\partial^2 u_x}{\partial x^2} + \frac{1-\mu}{2}\frac{\partial^2 u_x}{\partial y^2} + \frac{1+\mu}{2}\frac{\partial^2 u_y}{\partial x \partial y}\right) + f_x = 0 \quad (3a)$$

$$\frac{E}{1-\mu^2}\left(\frac{\partial^2 u_y}{\partial y^2} + \frac{1-\mu}{2}\frac{\partial^2 u_y}{\partial x^2} + \frac{1+\mu}{2}\frac{\partial^2 u_x}{\partial y \partial x}\right) + f_y = 0 \quad (3b)$$

where $x$ and $y$ denote the spatial coordinates, while $u_x, u_y$ represent the displacement components in the respective directions. The terms $f_x, f_y$ are the source terms, corresponding to body forces per unit volume. The material parameters include the Young's modulus $E$ and the Poisson's ratio $\mu$.

The solution mapping of the governing equations involves determining the target displacement field $(u_x, u_y)$, given the body force terms $f_x$, the displacement boundary conditions $g(x), x \in \partial\Omega$, and the material parameters $E$ and $\mu$. In our numerical experiments, we employ two types of data: data obtained from finite element (FEM) simulations, and DE-ruled data generation, in which random displacement fields are prescribed and the corresponding force terms are computed via the governing equations. As shown in Table 2, our model achieves high prediction accuracy on both datasets, demonstrating robust and consistent generalization performance across both scenarios.

## Wave equations

The wave equation is a second-order linear partial differential equation that governs the propagation of various physical wave phenomena in both space and time. In particular, the equation that characterizes the propagation of elastic waves within an isotropic, linear elastic medium is referred to as the elastic wave equation. Its specific form captures the dynamic behavior of mechanical disturbances as they travel through such media:

$$\rho \frac{\partial^2 u_i}{\partial t^2} = \frac{\partial}{\partial x_i}\left(\lambda \frac{\partial u_j}{\partial x_j}\right) + \frac{\partial}{\partial x_j}\left(\mu \left(\frac{\partial u_i}{\partial x_j} + \frac{\partial u_j}{\partial x_i}\right)\right) + f_i \tag{4}$$

where $u_j$ denotes the displacement response, and $f_i$ represents the source term, corresponding to the body force per unit volume—that is, the external excitation applied to the medium. The $\rho(x)$ denotes the mass density, while $\lambda(x)$ and $\mu(x)$ are the Lamé parameters. In geophysical applications, the Lamé parameters are typically derived from the shear wave velocity $C_s$ and the density $\rho(x)$.

The solution mapping of the elastic wave equation involves computing the time-evolving displacement response given the external source excitation $f_x$, displacement boundary conditions $g(x), x \in \partial\Omega$ and the initial condition $u_0(x)$. In this study, we implement our proposed AI solver on both FM generated dataset and DE-ruled generated dataset. For the DE-ruled data generation, we construct displacement fields with physically meaningful spatiotemporal structures that are consistent with elastic wave behavior, and then derive the corresponding source terms by substituting these fields into the DEs. For the FM simulation dataset, we model seismic excitation using the Ricker wavelet, which is a commonly used source function in seismology and elastic wave modeling:

$$F(t) = A_{Ricker}\left[1 - 2\pi^2 f_{Ricker}^2 (t - t_0)^2\right]$$

where $f_{Ricker}$ denotes the peak frequency of the Ricker wavelet, and $A_{Ricker}$ represents its amplitude. As shown in Table 2, our model effectively learns the solution across different data generation settings, achieving high prediction accuracy.

## Equation of motion

In the field of structural dynamics, the dynamic response of a multi-degree-of-freedom (MDOF) system subjected to external excitation is governed by a second-order nonlinear ordinary differential equation, commonly referred to as the equation of motion (EoM). Its general form is given by

$$\boldsymbol{M}\ddot{\boldsymbol{u}}(t) + \boldsymbol{C}\dot{\boldsymbol{u}}(t) + \boldsymbol{F}(t) = \boldsymbol{P}(t) \tag{5}$$

where $\boldsymbol{u}(t), \dot{\boldsymbol{u}}(t)$ and $\ddot{\boldsymbol{u}}(t)$ represent the displacement, velocity, and acceleration of the system, respectively. $M$ denotes the mass matrix of the multi-degree-of-freedom (MDOF) system, while $C$ is the viscous damping matrix; $\boldsymbol{F}_r(t)$ corresponds to the nonlinear restoring force of the system, and $P(t)$ represents the external load or excitation.

The Bouc–Wen hysteresis model[33] is a widely used mathematical model for representing nonlinear hysteretic behavior in dynamic systems. It has extensive application in structural engineering and has been further extended and adapted for use in various engineering problems:

$$\begin{aligned} F(t) &= \alpha k_i u(t) + (1-\alpha) k_i z(t) \\ \dot{z}(t) &= A\dot{u}(t) - \beta|\dot{u}(t)||z(t)|^{n-1}z(t) - \gamma \dot{u}(t)|z(t)|^n \end{aligned} \tag{6}$$

where $F_r(t)$ represents the restoring force, and $K$ is the initial elastic stiffness matrix of the system. The parameter $\alpha = k_f/k_i$ denotes the ratio of the post-yield stiffness $k_f$ to the pre-yield stiffness $k_i = F_y/u_y$, where $F_y$ is the yield force and $u_y$ is the yield displacement. The variable $z(t)$ is an unobservable internal hysteretic variable (commonly referred to as the hysteretic displacement). The dimensionless parameters $A, \beta > 0, \gamma, n$ govern the shape and behavior of the hysteresis loop, controlling the nonlinear characteristics of the model.

The MDOF Bouc–Wen model is formulated as a coupled system consisting of $N$ second-order differential equations (governing the primary motion) and $N$ first-order differential equations (governing the evolution of the hysteretic components). The solution mapping learned by our proposed AI solver aims to predict the time-dependent displacement response at each degree of freedom, given the external excitation and the constitutive parameters of the system. In our study, we adopt both numerical simulation data and DE-ruled data generation. For the numerical simulation data, we generate synthetic seismic load inputs to construct the excitation source, and solve the resulting motion equations using FM methods to obtain the displacement responses. For the DE-ruled data generation, we first prescribe displacement trajectories for the MDOF system, then solve the first-order differential equations in the Bouc–Wen restoring force model numerically to compute the hysteretic variables, and finally derive the corresponding external loads. As shown in Table 2, our proposed AI solver is capable of accurately learning the underlying dynamics for both datasets, demonstrating the robust performance across different data generation strategies.

**Forced Lorenz system**

The Lorenz system consists of three coupled first-order ordinary differential equations and is one of the most iconic examples in the theory of deterministic chaos[34]. A typical mathematical formulation of this externally driven system is

$$\begin{cases} \dot{x} = \sigma(y - x) \\ \dot{y} = \rho x - y - xz \\ \dot{z} = -\beta z + xy - f \end{cases} \quad (7)$$

where $x, y, z$ are the state variables, and $t$ represents time. The term $f$ denotes an external forcing function. The parameters $\sigma, \rho, \beta$ correspond to the Prandtl number, Rayleigh number, and a geometric factor, respectively, with $\beta$ commonly set to 8/3. Notably, for the parameter values $(\sigma, \beta) = (10, 8/3)$ and $\rho > \rho_H \approx 24.74$, the system exhibits sustained chaotic behavior.

For this equation, the model is required to learn the temporal evolution mapping from the forcing term to the system's state variables. We constructed two types of datasets for this problem. In the numerical simulation dataset, we apply a periodic forcing function, $f = A\sin(wt)$, along with perturbations to the initial conditions, to simulate the system dynamics. In the DE-ruled data generation, we first prescribe a basic form for the state variable $x$, then use the first equation of the system to solve for $y$, followed by the second equation to solve for $z$, and finally use the third equation to compute the corresponding forcing term $f$ (shown in Fig. 1). As shown in Table 2, the experimental results demonstrate that our proposed AI solver effectively learns the dynamics of the forced Lorenz system, achieving a high level of prediction accuracy.

# Conclusion

In this study, we propose a DE-ruled data generation method to address the scarcity of scientific training data for AI-based DE solvers. Additionally, we introduce a novel spatiotemporally coupled attention-based transformer architecture, integrated with SDO, designed to solve multiscale DEs. We verify both the data generation approach and the AI solver across a broad spectrum of DEs governing diverse planetary phenomena. The following conclusions can be drawn.

Within the DE-ruled data generation framework, prior knowledge or Gaussian random fields can be employed to synthesize spatiotemporal data representing target solutions. These generated solutions are then substituted into the DEs to derive corresponding source terms and initial/boundary conditions, thereby constructing well-posed equation–solution data pairs. This approach enables efficient, low-cost computation of arbitrarily large first-principles-adherent datasets for training AI solvers. Furthermore, to enhance AI solvers' ability to resolve high-wavenumber components, we introduce a reversible SDO based on Fourier transform. This operator maps high-wavenumber features to a low-wavenumber space. Leveraging this capability, we develop a spatiotemporally coupled attention-based transformer architecture integrated with the SDO for solving multiscale DEs. We derive the upper bound of the Hessian condition number of the loss function with respect to the spatial gradient magnitude of the target solution data, demonstrating a smoother loss landscape after scale dilation.

We thoroughly validated our proposed methods across a wide range of scientific phenomena governed by three PDEs and two ODEs, consistently achieving high accuracy. Our approach yielded particularly compelling results for the highly challenging multiscale Navier-Stokes equations, exhibiting superior relative error performance compared to state-of-the-art methods such as FNO, U-NO, Transolver, and LSM. These findings unequivocally verify the efficacy of both our DE-ruled data generation method and our AI solver in learning physically consistent solutions.

## Methods

**Differential equation-ruled data generation**

We present a low-cost data generation method that satisfies the governing physical constraints. For physical quantities with well-established prior knowledge, e.g., turbulence, the spatial energy distribution can be readily obtained based on turbulence energy spectra derived from Kolmogorov's K41 theory[27] or experimental observations, such as the widely accepted spectrum proposed by Pope[26]. By incorporating temporal evolution on this spatial structure, a spatiotemporal velocity field can then be constructed[28]. Then, the divergence-free velocity can be generated as

$$u_i = 2 \sum_{m=1}^{M} q_m \cos(\boldsymbol{K}_m \boldsymbol{X} - \psi_m) \sigma_{i,m} \tag{8}$$

where $\boldsymbol{K}_m \in [0, W] \times \boldsymbol{k}$ and $\boldsymbol{X} \in [0, T] \times \Omega$ denote the generalized wavevector of the $m$-th mode in Fourier space and the generalized spatial-temporal position in the physical domain, respectively. The $\psi_m$ represents a random phase associated with the $m$-th mode, while $\sigma_{i,m}$ is a unit vector orthogonal to the corresponding wavevector, ensuring divergence-free constraint. The coefficient $q_m$ encodes the energy content of the $m$-th mode according to the prescribed energy spectrum.

Once the velocity field solution is constructed, all terms in the governing PDE—including the initial and boundary conditions—are explicitly defined. The corresponding source term $f(t, \boldsymbol{x})$ can then be computed directly by applying the temporal derivative operator $\mathcal{L}_t$ and the spatial (linear or nonlinear) differential operator $\mathcal{L}_x$ as specified in equation DE (1), thereby ensuring that the generated data strictly satisfies the governing equations. Taking the incompressible Navier–Stokes equation (2) as a representative case (the approach similarly applies to other types of PDEs or ODEs), we begin with a prescribed velocity field $u_i(t, \boldsymbol{x})$. At each time instant, the corresponding pressure field $p(t, \boldsymbol{x})$ is recovered by solving the pressure Poisson equation: $\nabla^2 p = -\left(\frac{\partial}{\partial x_i}\left(u_j \frac{\partial u_i}{\partial x_j}\right)\right)$. Once both the velocity and pressure fields are available, the remaining terms in the momentum equation are evaluated in sequence: the temporal derivative $\partial u_i / \partial t$, the convective term $u_j \partial u_i / \partial x_j$, the pressure gradient $\partial p / \partial x_i$, and the viscous diffusion term $Re^{-1} \nabla^2 u_i$. The residual source term $f(t, \boldsymbol{x})$ is then computed as the quantity required to balance all terms in the equation.

**Theorem 1**. Consider a spatiotemporally coupled attention-based transformer architecture learning the solution $u \in \mathbb{R}^d$ of differential equations, with source term and initial/boundary conditions $(f, u_{t_0}, g)$ as input, and outputs $u_{t_1}, u_{t_2}, \ldots, u_{t_n}$. If the Jacobian of the loss function w.r.t. the model parameters has full column rank, then the condition number of the Gauss-Newton matrix, $\kappa(H_{GN})$, is bounded by:

$$\kappa(H_{GN}) \leq \frac{C_J^2 S^2}{\tilde{\lambda}}$$

where $\kappa$ denotes the condition number, $H_{GN}$ is the Gauss-Newton approximation of the neural network's loss function Hessian matrix, $C_J$ is a constant related to the neural network's layer-wise Lipschitz constant, $S$ is the maximum magnitude of the solution and source term's spatial gradient, and $\tilde{\lambda} > 0$ is the minimum eigenvalue of $H_{GN}$.

As shown in Theorem 1, a smaller gradient magnitude of the solution directly contributes to a lower condition number of the loss function's Hessian matrix, leading to a smoother, more trainable loss landscape.

# Acknowledgments

The authors would like to acknowledge the National Natural Science Foundation of China (Grant Nos. 52441803, 52441802, 12472289 and 92152301).

# Data availability

The dataset generation scripts used in this study will be publicly available in a GitHub repository.

# Code availability

The code used in this study will be released in a GitHub repository.

# Competing interests

The authors declare no competing interests.